\documentclass{article}
\usepackage{ijcai23}
\usepackage{listings}
\usepackage{times}
\usepackage{soul}
\usepackage{url}
\usepackage[hidelinks]{hyperref}
\usepackage[utf8]{inputenc}
\usepackage[small]{caption}
\usepackage{graphicx}
\usepackage{amsmath}
\usepackage{amsthm}
\usepackage{booktabs}
\usepackage{algorithm}
\usepackage{algorithmic}
\usepackage[switch]{lineno}
\usepackage{amsfonts}
\usepackage{tcolorbox}

\title{Integrating Stock Features and Global Information via Large Language Models for Enhanced Stock Return Prediction} 

\author{
Yujie Ding\thanks{Equal contribution.} 
\and
Shuai Jia\footnotemark[1]
\and
Tianyi Ma\thanks{All correspondence to: matianyi@myhexin.com} \and
Bingcheng Mao\and
Xiuze Zhou\and
Liuliu Li\And
Dongming Han
\affiliations
Hithink RoyalFlush Information Network Co., Ltd.\\
\emails
\{dingyujie2, jiashuai, matianyi, maobingcheng, zhouxiuze, liliuliu, handongming2\}@myhexin.com
}

\begin{document}

\maketitle

\begin{abstract}
The remarkable achievements and rapid advancements of Large Language Models (LLMs) such as ChatGPT and GPT-4 have showcased their immense potential in quantitative investment. Traders can effectively leverage these LLMs to analyze financial news and predict stock returns accurately. However, integrating LLMs into existing quantitative models presents two primary challenges: the insufficient utilization of semantic information embedded within LLMs and the difficulties in aligning the latent information within LLMs with pre-existing quantitative stock features. We propose a novel framework consisting of two components to surmount these challenges. The first component, the Local-Global (LG) model, introduces three distinct strategies for modeling global information. These approaches are grounded respectively on stock features, the capabilities of LLMs, and a hybrid method combining the two paradigms. The second component, Self-Correlated Reinforcement Learning (SCRL), focuses on aligning the embeddings of financial news generated by LLMs with stock features within the same semantic space. By implementing our framework, we have demonstrated superior performance in Rank Information Coefficient and returns, particularly compared to models relying only on stock features in the China A-share market. 
\end{abstract}

\section{Introduction}

With the continuous expansion of stock market capitalization, trading stocks has garnered significant interest among investors as an appealing investment option. Consequently, the field of stock return prediction has gained considerable attention~\cite{abe2020cross}. The primary objective of stock return prediction is to assist investors in making informed investment decisions by forecasting the returns of stocks. To effectively capture the essence of predictive stock returns, numerous factors with strong explanatory power have been proposed~\cite{yogo2006consumption,nakagawa2020ric}. In the intricate and dynamic stock market environment, staying informed about relevant financial news is crucial for investors seeking to make well-informed decisions. However, many investors struggle to extract key insights from numerous financial news when confronted with vast amounts of textual data.

Recently, Large Language Models (LLMs), such as BERT-based \cite{devlin2018bert,liu2019roberta}, and GPT-based models \cite{brown2020language}, have demonstrated exceptional success in a wide range of Natural Language Processing (NLP) tasks. These models leverage contextualized representations and extract valuable information from vast text corpora. In the field of finance, researchers have begun exploring the application of LLMs to enhance decision-making processes. For instance, Lopez-Lira et al. \cite{lopez2023can} employ ChatGPT to perform sentiment analysis on news headlines, thereby improving decision-making in stock trading. Xie et al.~\cite{xie2023wall} have devised a range of prompting strategies to enhance the financial analysis capabilities of ChatGPT. However, such approaches face the following significant challenges. Firstly, they rely solely on generated sentiment information or inherent reasoning abilities of LLMs and do not delve deeper into other valuable knowledge in LLMs. Consequently, valuable information beyond sentiment is overlooked, potentially limiting the model's ability to make more informed decisions. Secondly, financial news can be categorized into three levels, i.e., macro (e.g., markets, policies, economy), meso (e.g., industries), and micro (e.g., stocks, companies), each offering distinct perspectives. These different levels pose difficulties in aligning and mapping their information into the same level directly, which presents a challenge when leveraging LLMs to incorporate a comprehensive understanding of the stock return prediction. An alternative solution is extracting news embeddings through LLMs and concatenating them with stock features. However, the embeddings generated by LLMs and the stock features do not inherently share the same modality and are not aligned within a unified semantic space. These issues highlight the need for further exploration and development in utilizing LLMs for stock return prediction. 

The contribution of this paper is summarized as follows. We propose a model-free framework called Self-Correlated Reinforcement Learning with the Local-Global model (SCRL-LG) to tackle the above-mentioned challenges. This framework harnesses the power of LLMs and explores their potential in multi-modal stock return prediction. We introduce a Local-Global (LG) model that decomposes stock returns into two distinct components. Firstly, we consider the idiosyncratic return inherent in the stock volume-price features. Secondly, we examine the features that affect stock returns, as reflected in market-related news, macroeconomic policies, industry trends, and specific stocks. Separating these components gives us a more nuanced understanding of the features contributing to stock performance. Furthermore, to improve the accuracy of stock return predictions, we align the information derived from stock features and news embeddings generated by LLMs within the same semantic space. This is achieved through the introduction of Self-Correlated Reinforcement Learning (SCRL). By aligning these disparate sources of information, we enable a more comprehensive and integrated understanding of the features that impact stock performance.

\section{Models}

\subsection{The Local-Global Model}\label{lgm}
Inspired by classical asset pricing models~\cite{jensen1972capital,fama1993common,fama2015five}, we decompose the stock return prediction model into two sub-models: a Local model and a Global model. The Local model models stock-specific, intrinsic information (such as volume, price, and other technical features) to predict stock returns, corresponding to the $\alpha$ component in asset pricing models. On the other hand, the Global model captures the global information, i.e., the impact of markets, industries, and policies on stock returns, corresponding to the $\beta$ component in asset pricing models. The final return of stock $i$ at time $t$, denoted as $r_{t, i}$, is defined as a combination of the Local and Global components:
\begin{equation} \label{camp}
    r_{t,i} = \alpha_{t, i} + \beta_{t, i} f_{t}, 
\end{equation}
where $f_{t}$ represents the function that models the alignments between stock features and the global information at time $t$. We denote the Local and Global components as follows: 
\begin{align}
    \alpha_t = F_{local}(M_{t-1}) \in R^{n_{t-1}, 1}, \\
    \beta_t = F_{beta}(M_{t-1}) \in R^{n_{t-1}, D}, \\
    f_t \in R^{D, 1}. 
\end{align}

Here, $F_{local}(M_{t-1})$ and $F_{beta}(M_{t-1})$ represent the models that predict the $\alpha$ and $\beta$ components, respectively. $D$ is the desired dimensionality of the aggregated vector (i.e., the output dimension of $f_{t}$). Please note that $F_{local}(M_{t-1})$, $F_{beta}(M_{t-1})$, and $f_t$ can be implemented using machine learning or deep learning models, such as Multi-Layer Perceptron (MLP). Based on the above descriptions, our proposed Local-Global model that estimates stock returns at time $t$ is defined as follows:
\begin{equation} \label{finpred}
     \hat{r_{t}} = F_{local}(M_{t-1}) + \underbrace{F_{beta}(M_{t-1}) \cdot f_t}_{F_{global}(\cdot)}.   
\end{equation}
We aim to minimize the loss $l$ between the actual returns $r_t$ and the estimated returns $\hat{r_{t}}$ (e.g., the MSE loss), where $r_t$ is a vector composed of $\{r_{t, i} \}_{i = 1}^{n_t}$ at time $t$. The critical part of this model lies in modeling $F_{global}(\cdot)$. Next, we present three different models for constructing the global model. 

\textbf{\textit{Model 1: Global model using only stock features}}. Inspired by the work of \cite{duan2022factorvae}, we utilize an attention mechanism to aggregate the stock features $M_t$ into a vector $f_t$, which captures global information that influences stock returns. First, we define two transformation matrices $W_{\text{key}}$ and $W_{\text{value}}$ of dimensions $m \times D$. We then compute the key matrix $K$ and the value matrix $V$ as follows:
\begin{equation}
    \begin{aligned}
    K = M_{t-1} \cdot W_{\text{key}}\in \mathbb{R}^{n_{t- 1}\times D}, \
    V = M_{t-1} \cdot W_{\text{value}}\in \mathbb{R}^{n_{t-1}\times D}.
    \end{aligned} 
\end{equation}

Next, we introduce a query vector $q\in \mathbb{R}^{1\times D}$ and compute the attention weights $a_{att}$ using the dot product between the query vector and the key matrix, normalized by their Euclidean norms:

\begin{equation}
    \begin{aligned}
    a_{att} = \max \left(0, \frac{q \cdot K^T}{\|q\|_{2} \cdot\|K\|_{2}} \right),
    \end{aligned} 
\end{equation}
where the max function ensures that the attention weights remain non-negative. We then normalize the attention weights by dividing them by their sum to ensure that the attention weights sum up to 1:
\begin{equation}
    \begin{aligned}
    a_{att} = \frac{a_{att}}{\text{sum}(a_{att})}.
    \end{aligned} 
\end{equation}

Finally, the alignment function of Model 1 is defined as follows:
\begin{equation}\label{features_only}
    \begin{aligned}
    f_t^{\text{stock}} = (a_{att} \cdot V)^{T} \in \mathbb{R}^{D\times 1}.
    \end{aligned} 
\end{equation}

However, both the global and local models have only utilized stock features. In this case, global information can be considered a transformation or compression of stock features. The two have redundant information, which may lead to overfitting. 

\label{lglobal}
\textbf{\textit{Model 2: Global model using only LLM}}. 
Stock return prediction using text-based methods presents challenges, and the prevailing approach involves employing text embeddings as features for individual stocks \cite{hu2018listening,chen2018leveraging}. These embeddings are then concatenated with the original feature set to facilitate prediction. However, much stock-related news has consisted of descriptions regarding specific price features. For example, ``Today, shares of {a company} were heavily bought, resulting in a 6\% increase in stock price". Such descriptions of stock prices have already overlapped with the existing features in the feature set. Consequently, we attempted to incorporate news on markets, industries, and government policies as prompts into LLMs to obtain embeddings, which were then used to construct global information. This type of information, such as ``Ministry of Commerce: Strengthening Business Environment Protection," is not directly related to specific stock features. It is expected that incorporating such information would capture broader market trends, industry developments, and regulatory changes that may impact the performance of individual stocks. This integration of global information into the model allows for a more comprehensive analysis, potentially uncovering hidden patterns and improving the accuracy of predictions. 

Now we define $f_t = F_{global}(V_{llm})$, where $V_{llm} \in \mathbb{R}^{1, d_{llm}}$ is the output vector of LLM when predicting the next token given the appended daily news as the input, and $d_{llm}$ is the dimension size. Here is an example of the inputs:
\begin{tcolorbox}[width=0.48\textwidth]
\textbf{Example of Prompt}:

China implements a fishing ban period system in key water areas of the Yellow River Basin

Local development of culture, finance, nighttime economy, intellectual property protection, and others will receive priority support

Thailand has reported 63 cases of Omicron variant importation

...

\{All the news headlines of the day\}
\end{tcolorbox}

Each day we generate one $V_{llm}$. It is worth noting that we have also explored various strategies for constructing few-shot prompts. For instance, we attempted to incorporate instructions such as analyzing market trends, predicting stock returns, and labeling each piece of information with industry and market-related tags. However, calculating the correlation between the daily $V_{llm}$ generated by these different prompt strategies revealed a remarkably high correlation. Consequently, the performance was unsatisfactory. On the other hand, the prompt construction method we proposed exhibited a lower correlation and yielded superior results. This phenomenon suggests that when the correlation between daily $V_{llm}$ is lower, there is increased information diversity, which is expected to supplement more global information. 

To align the stock features and $V_{llm}$, we define parameters $W_{llm} \in \mathbb{R}^{D, d_{llm}}$ that can be learned by minimizing the loss function $l$, and the alignment function of Model 2 becomes
\begin{equation} \label{only_LLM}
    \begin{aligned}
        f_t^{llm} = W_{llm} V_{llm}^T \in \mathbb{R}^{D, 1}. 
    \end{aligned}
\end{equation}

However, Model 2 may not effectively align stock features and $V_{llm}$ generated by LLMs. The dimensionality of $W_{llm}$ could be significant ($D \times d_{llm}$), and such overparameterization can lead to overfitting. 

\textbf{\textit{Model 3: Global model with both stock features and LLM}}. 
Based on the issues identified in Model 1 and Model 2, we propose incorporating the learning of a sparse vector $V_{sparse}$ to align stock features and $V_{llm}$, serving as a feature selector or a dropout variant on features to mitigate overfitting. We outline the following method for generating global information:
\begin{equation} \label{eq:sparseG}
f_t = f_t^{\text{stock}} \odot V_{sparse},
\end{equation}
where $f_t^{\text{stock}}$ is the alignment function defined in Model 1, and $V_{Sparse}$ is a sparse vector (e.g., $\begin{bmatrix}0,0,1,\cdots,1, 0 \end{bmatrix}$) transformed from $f_t^{llm}$. The transforming processes will be presented in section \ref{sec:SCRL}. The element-wise multiplication ($\odot$) is performed to incorporate the sparse information into the global representation. Now, the complete estimation of stock returns $\hat{r_t}$ becomes
\begin{equation} \label{camp_full}
    \hat{r_t} = F_{local}(M_{t-1}) \\
    + F_{beta}(M_{t-1}) \\
    \cdot (f_t^{stock} \odot V_{sparse}).
\end{equation}

Here, $V_{sparse}$ has the same dimension size as the stock features, and it can be used as a feature selector to constrain the generation of $f_t$ from stock features to reduce redundancy, mitigate overfitting, and improve model performance. However, the semantic space of the $V_{llm}$ generated by LLMs differs from that of stock features. Additionally, learning $V_{sparse}$ may lead to the problem of vanishing gradients. In this context, we propose a novel method to align different data effectively.

\begin{figure}[h]
   \centering
   \includegraphics[width=1\linewidth]{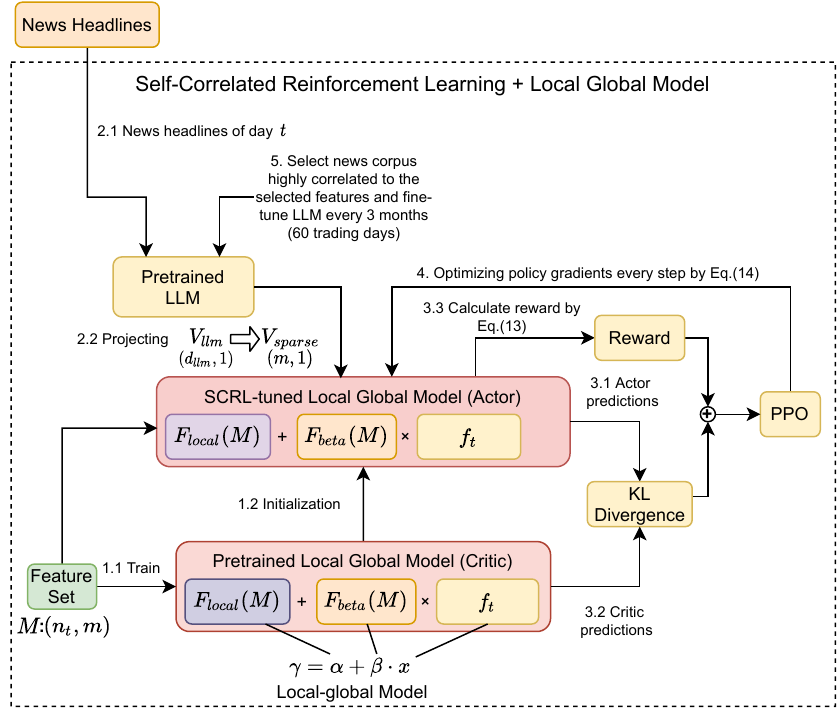}
   \caption{The Self-Correlated Reinforcement Learning with Local Global Model. }
   \label{fig:method}
\end{figure}

\subsection{Self-Correlated Reinforcement Learning}\label{sec:SCRL}

We propose a Self-Correlated Reinforcement Learning (SCRL) framework, incorporating the Proximal Policy Optimization (PPO) paradigm \cite{ppo}, to align the stock features with the global information embedded in LLM embeddings within the same semantic space. The complete processes of SCRL with the Local-Global model are illustrated in Figure \ref{fig:method}. To extract the embeddings of news effectively, we pre-train an LLM (e.g., Llama \cite{touvron2023llama}) with a curated news corpus, providing a rich representation superior to traditional handcrafted feature extraction methods. 

Subsequently, we train the Local-Global model (the Critic Model) using a supervised learning approach with a stock feature set $M$ (Step 1.1 in Figure \ref{fig:method}). Following this, we initialize the Actor model using the trained Critic model (Step 1.2 in Figure \ref{fig:method}) and fine-tune the Actor model. This training mode allows the model to learn more complex decision boundaries, thus improving predictive performance.

To accomplish this, we utilize the pre-trained LLM to generate $V_{llm}$ given the appended news headlines of the day $t$ (Step 2.1). Subsequently, we employ a mapping module (a linear layer) to map $V_{llm}$ in $f_t^{llm}$ to the same dimension as the stock feature set, obtaining the feature vector $V_{sparse}$ (Step 2.2). This enables the integration of information from different sources into the same semantic space, thereby improving predictive accuracy.

Next, we generate probability distributions using the Critic and Actor models (steps 3.1-3.2). Then, we calculate the overall reward function (step 3.3), which is defined as follows: 
\begin{equation} \label{kl}
    R(x,y)=r(x,y)-\theta KL(x,y), 
\end{equation}
where $R(x, y)$ represents the overall reward function, $r(x, y)$ is the result of the reward model, $\theta$ is a hyper-parameter regulating the balance between the reward and the KL divergence, and $KL(x,y)$ is the KL divergence between the predictions generated by the Actor model and the Critic model to ensure that our trained distribution remains close to the original. Specifically,  $KL(x,y)$ is defined as
\begin{equation} \label{kl-app}
    KL(x,y) = \log \left(\pi_\phi^{\mathrm{SR}}(y \mid x) / \pi^{\mathrm{REF}}(y \mid x)\right),
\end{equation}
where $\pi_\phi^{\mathrm{SR}}$ is the probability of the Actor model predicting output $y$ given input $x$. $\pi_\phi^{\mathrm{REF}}$ is the prediction probability of the Critic model. By generating probability distributions, we create a decision-making process that can select investment strategies in an optimized manner. The defined reward function $R(x, y)$ incentivizes the model to balance the reward and the KL divergence, leading to an efficient and stable policy. 

We then optimize the model using PPO with fixed-length trajectory segments. Each $N$ participant collects data for $T$ time steps, and we build a proxy loss on $N_t$ time steps and optimize it using mini-batch SGD (Step 4), which effectively deals with the issues of sample efficiency and exploration-exploitation trade-off in reinforcement learning. Fixed-length trajectory segments allow the model to consider future rewards at each step, which is crucial for sequential decision-making problems.

Finally, we select the news related to the selected stock features from the previous 60 trading days in the news corpus. We append the selected news on the training corpus, retrain the LLM (Step 5), which is meant to help the model to continually adapt to news and market, and repeat steps 1-5 until the Actor model converges. 

\section{Experiments}
We conduct empirical experiments to validate the performance of the proposed models. We first describe the datasets and model settings, then analyze the experiment results.  

\subsection{Settings}
\textbf{\textit{Datasets}}. The experiments are conducted on the China A-shares market using a dataset comprising $3506$ stocks and $342$ daily features constructed from stock price-volume data ranging from $01/01/2019$ to $12/31/2022$. We exclude suspended stocks or the stocks listed after $2020$. Additionally, we collect news headlines during the same period, resulting in a dataset size of $162,845$ headlines and $2,526,174$ stock data points for training (from $01/01/2019$ to $12/31/2021$) and $849,173$ data points for testing. 

\textbf{\textit{Models}}. We compare the proposed SCRL-LG model described in Eq.\eqref{camp_full} with three models mentioned in Section \ref{lgm}:
\begin{itemize}
\item \textbf{Local (w/o Global model)}: This model applies only the $F_{\text{local}}(M_{t-1})$ part of Eq.\eqref{camp_full}.
\item \textbf{LG-STOCK (w/o LLM in Global model)}: This Local-Global model uses only stock features for the global model, as described in Eq.\eqref{features_only}.
\item \textbf{LG-LLM (w/o stock features in Global model)}: This Local-Global model uses only representations generated by LLM for the global model, as described in Eq.\eqref{only_LLM}.
\end{itemize}

We utilize a two-layer neural network with the Rectified Linear Unit (ReLU) as the activation function for $F_{\text{local}}(\cdot)$ and $F_{\beta}$. The network configuration is $[324, 36, 1]$. In the PPO part, we use $2,048$ steps, a learning rate of $0.00025$, a batch size of $128$, and a reward scaling factor of $1 \times 10^{-4}$.

\textbf{\textit{LLM}}. We utilize Llama (7B) \cite{touvron2023llama}, an efficient open LLM. To enhance the Chinese generation capabilities of the Llama model, we continued training Llama for an additional 2 epochs on 15 GB of Chinese corpus, specifically the CLUECorpusSmall \cite{CLUECorpus2020} and the Chinese Scientific Literature Dataset \cite{li-etal-2022-csl}. These datasets encompass a wide range of Chinese text sources, including news articles, web content, Wikipedia entries, comments, and scientific literature, from 2010 to 2020. During the training process, we utilized the Chinese tokenizer employed by Cui et al. \cite{cui2023efficient} and expanded the vocabulary size from 32,000 to 49,953. The Llama model's embeddings were expanded to accommodate this enlarged vocabulary size. The batch size in tokens was 128K, while the learning rate was 2e-5. Additionally, we applied cosine decay as the learning rate scheduler. We trained and fine-tuned the model using 8 Nvidia A100 GPUs (80 GB). It is worth mentioning that while the training corpus for the original Llama may include samples extracted from Wikipedia in 2022, these samples account for approximately 4.5\% of the total training data, and their potential impact on information leakage can be considered negligible. 

\textbf{\textit{Backtesting}}. To mitigate potential information leakage from the training corpus of Llama, we conducted a backtest from $01/01/2022$ to $12/31/2022$, following a similar approach proposed in \cite{lopez2023can}. We conducted daily sorting of stocks based on predicted returns and divided them into ten quantiles. Our strategy involved purchasing an equal allocation of stocks from the top 10\% quantile while liquidating portfolios not belonging to this quantile. All transactions were executed at the closing prices, with a transaction cost of 0.3\%. Additionally, we restrict our focus to news released between the end of the previous trading day at 3:30 PM and the commencement of the current trading day at 9:30 AM. 

\textbf{\textit{Evaluation Metrics}}. We evaluate the Rank Information Coefficient (Rank IC), a widely used financial ranking metric, for all stocks. We also evaluate the top 10\% quantile stocks' cumulative returns, annual returns, Sharpe ratio, max drawdown (MDD), and turnover rates. 

\begin{figure}[t]
   \centering
   \includegraphics[width=1\linewidth]{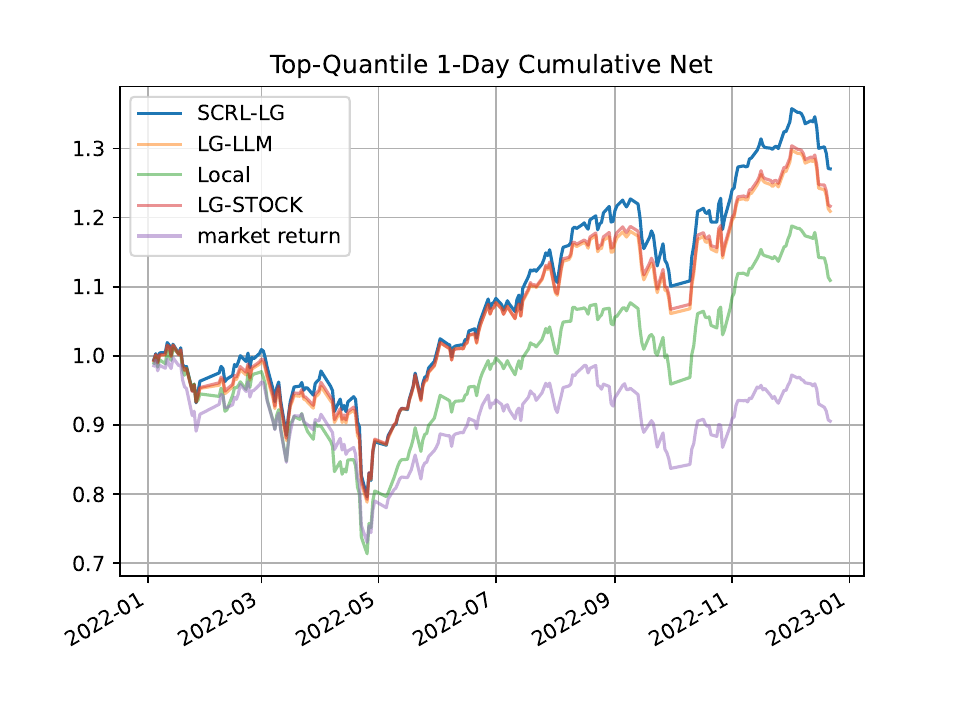}
   \caption{The cumulative return of models. }
   \label{fig:cumret}
\end{figure}

\begin{table*}[t]
  \centering
  \caption{Model performance and standard deviation. }
    \begin{tabular}{l|ccccc}
    \toprule
                               & RankIC        & Annual Return  & Top Minus Bottom & Sharpe Ratio & MDD \\ \hline
    Local                     & $0.132\pm0.002$  & $0.112\pm0.02$ & $1.14\pm0.08$  & $0.56\pm0.07$ & $0.297\pm0.01$ \\
    LG-STOCK & $0.142\pm0.004$  & $0.216\pm0.03$ & $1.67\pm0.12$ & $1.01\pm0.12$ & $0.223\pm0.02$ \\
    LG-LLM    & $0.144\pm0.003$  & $0.209\pm0.04$ & $1.63\pm0.10$ & $0.98\pm0.11$ & $0.227\pm0.01$ \\
    SCRL-LG   & $0.152\pm0.003$  & $0.288\pm0.05$ & $1.83\pm0.11$ & $1.24\pm0.15$ & $0.219\pm0.02$ \\
    \bottomrule
    \end{tabular}
    \label{ablation}
\end{table*}

\begin{figure*}[t]
   \centering
   \includegraphics[width=1\linewidth]{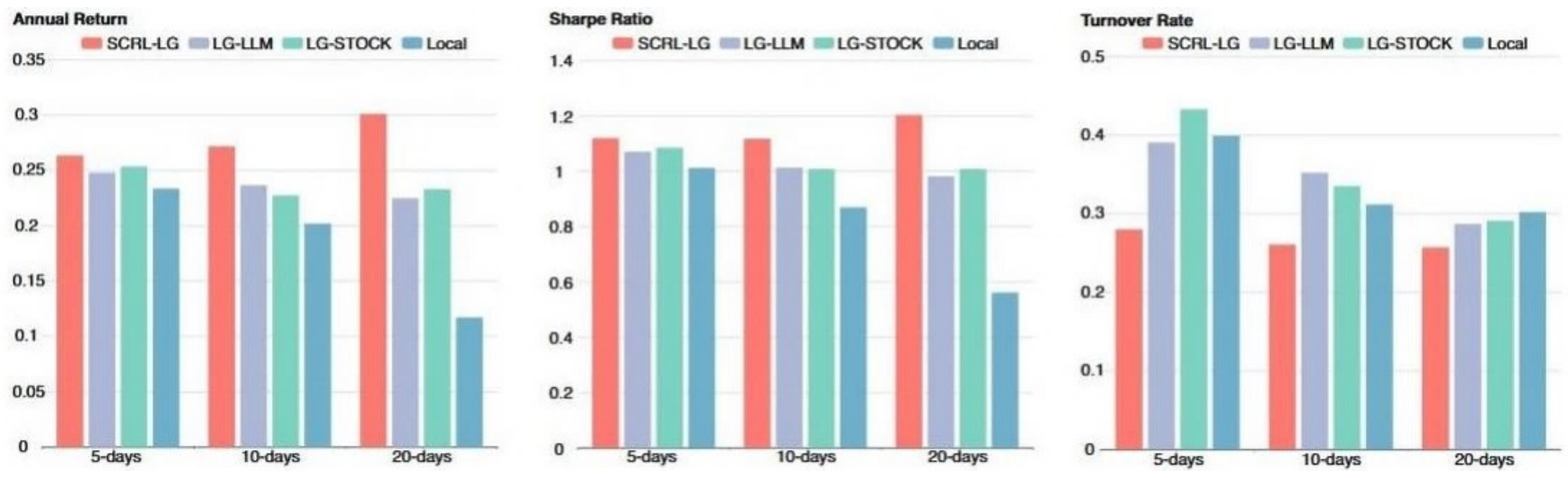}
   \caption{The model performance when predicting stock returns of the next 5, 10, and 20 trading days. }
   \label{fig:days}
\end{figure*}

\subsection{Results}

Firstly, we evaluate the profitability of models. Figure \ref{fig:cumret} and Table \ref{ablation} present models' performance in different metrics. From the experimental results, it can be observed that the Local model performs the worst. In comparison, the two methods that combine global information, namely LG-STOCK and LG-LLM, exhibit significant improvements in all metrics. This suggests that 1) predicting stock returns based on feature-generated global information is effective, and 2) the information generated by LLMs can significantly enhance predictive performance. Finally, the proposed method aligns the two types of global information through SCRL and combines the strengths of LG-STOCK and LG-LLM. The results show that SCRL-LG yields noticeable improvements in all metrics compared to the methods above for generating global information. This indicates that the method proposed in this paper is more effective in integrating financial news information extracted by LLM with the current stock features, thereby enhancing the predictive capabilities of existing features and models.

We assess the predictive accuracy of LLM for stock returns at different time horizons, as well as the impact of transaction costs. We trained three models to predict stock returns for 5, 10, and 20 trading days ahead. Figure \ref{fig:days} presents these three models' annual returns, Sharpe ratios, and turnover rates. From Figure \ref{fig:days}, it can be observed that as the prediction horizon increases, the performance gaps between the different methods become more pronounced. Based on empirical observations, shorter-term return predictions (e.g., 5 days) exhibit greater volatility due to market noise, resulting in relatively smaller performance discrepancies among the various methods. On the other hand, longer-term return predictions (e.g., 20 days) tend to average out the noise over time, resulting in reduced noise levels and widening performance gaps between different methods. At this point, the effectiveness of incorporating various global information approaches gradually becomes evident.

Generally, shorter-term prediction models are expected to perform better under a 1-day holding period. From Figure \ref{fig:days}, it can be seen that the backtesting results of the Local model display a noticeable decrease in annual returns and Sharpe ratio as the prediction horizon extends. Conversely, the models incorporating global information (LG-STOCK and LG-LLM) demonstrate relatively minor declines in performance. Surprisingly, the SCRL-LG model even exhibits performance improvements with increasing prediction horizons, accompanied by a decrease in turnover rate, thus confirming the effectiveness of this method. These results collectively indicate that the three proposed approaches for incorporating global information can effectively mitigate the decline in predictive performance observed in the Local model.

\section{Related Work}

Over the past few decades, numerous single-factor and multi-factor asset pricing models have been proposed and are widely used in predicting stock returns \cite{jensen1972capital,fama1993common,fama2015five}. In recent years, asset pricing methods based on deep learning have emerged \cite{gu2020empirical,giglio2022factor,duan2022factorvae,chen2023deep}. These methods adhere to the framework of traditional asset pricing models and incorporate deep learning techniques to synthesize factors. However, this fusion reduces the interpretability of the factors, while emphasizing the validation of their out-of-sample performance.

With the rising popularity of Large Language Models (LLMs), researchers have begun to explore their application in stock return prediction. There are two main categories of research in this area. The first category uses historical stock price data with LLMs to predict stock returns. Essentially, this type of research tackles sequence prediction problems. In \cite{gupta2022hisa}, an LLM is used as an emotion recognition component. It analyzes Twitter public opinion data and generates a score, which, together with stock price data, is input into a Gated Recurrent Unit (GRU) model for stock return prediction.

However, Xie et al. ~\cite{xie2023wall} suggest that ChatGPT performs poorly in multivariate stock return prediction tasks, underperforming even basic methods such as linear regression. Nonetheless, prompting strategies, such as chains of thought, have proven effective. The work of Wu et al. \cite{wu2022s_i_lstm} shares a similar fundamental concept with that of Gupta et al. \cite{gupta2022hisa}, but it adopts a more rudimentary sentiment analysis using traditional methods like word2vec and CNN instead of an LLM. Additionally, it incorporates technical factor data into the training dataset. In summary, the effectiveness of LLMs in predicting stock returns from sequential data is not exceptional. However, the inclusion of critical data, such as factor data, or the use of appropriate prompting strategies like the Chain of Thoughts (CoT), has proven effective.

The second research category directly uses LLMs to score stocks by leveraging the text data that LLMs excel at processing. The work of Lopez et al. \cite{lopez2023can} and Ko et al. \cite{ko2023can} exemplifies this approach. Ko et al. \cite{ko2023can} employed a straightforward operation without data fine-tuning to create a candidate set of stocks. The model is prompted to behave as a professional stockbroker and selects stocks from this set. The results indicate that this approach outperforms random selection. In \cite{lopez2023can}, news headline data is used to prompt whether a specific piece of information is advantageous for a stock, yielding three possible outcomes: ``yes," ``no," or ``uncertain." The final decision is based on the cumulative scores from all the prompts. The article claims a 500\% gain from this approach.

\section{Conclusions}
To demonstrate the effectiveness of LLMs in predicting stock returns, we have introduced a novel approach known as Self-Correlated Reinforcement Learning with Local-Global model (SCRL-LG). Within the SCRL-LG framework, LLMs act as stock feature selectors, extracting meaningful representations from news headlines. These extracted representations are subsequently employed in RL to acquire precise feature alignments. Through meticulous experimentation conducted on China A-share data from 2022, our LLMs-based approach has demonstrated notable advantages and significantly improved the accuracy of stock return predictions. In our ongoing pursuit of advancement, our foremost objective is to enhance the representation capabilities of LLMs. We aim to refine the underlying mechanisms of LLMs to optimize their predictive accuracy further. Furthermore, we are committed to exploring the potential of LLMs in predicting trends across various domains such as markets, macroeconomics, and industries. 

\bibliographystyle{named}

\bibliography{ijcai23}

\begin{thebibliography}{}

\bibitem[\protect\citeauthoryear{Abe and Nakagawa}{2020}]{abe2020cross}
Masaya Abe and Kei Nakagawa.
\newblock Cross-sectional stock price prediction using deep learning for actual
  investment management.
\newblock In {\em Proceedings of the 2020 Asia Service Sciences and Software
  Engineering Conference}, pages 9--15, 2020.

\bibitem[\protect\citeauthoryear{Brown \bgroup \em et al.\egroup
  }{2020}]{brown2020language}
Tom Brown, Benjamin Mann, Nick Ryder, Melanie Subbiah, Jared~D Kaplan, Prafulla
  Dhariwal, Arvind Neelakantan, Pranav Shyam, Girish Sastry, Amanda Askell,
  et~al.
\newblock Language models are few-shot learners.
\newblock {\em Advances in neural information processing systems},
  33:1877--1901, 2020.

\bibitem[\protect\citeauthoryear{Chen \bgroup \em et al.\egroup
  }{2018}]{chen2018leveraging}
Weiling Chen, Chai~Kiat Yeo, Chiew~Tong Lau, and Bu~Sung Lee.
\newblock Leveraging social media news to predict stock index movement using
  rnn-boost.
\newblock {\em Data \& Knowledge Engineering}, 118:14--24, 2018.

\bibitem[\protect\citeauthoryear{Chen \bgroup \em et al.\egroup
  }{2023}]{chen2023deep}
Luyang Chen, Markus Pelger, and Jason Zhu.
\newblock Deep learning in asset pricing.
\newblock {\em Management Science}, 2023.

\bibitem[\protect\citeauthoryear{Cui \bgroup \em et al.\egroup
  }{2023}]{cui2023efficient}
Yiming Cui, Ziqing Yang, and Xin Yao.
\newblock Efficient and effective text encoding for chinese llama and alpaca.
\newblock {\em arXiv preprint arXiv:2304.08177}, 2023.

\bibitem[\protect\citeauthoryear{Devlin \bgroup \em et al.\egroup
  }{2018}]{devlin2018bert}
Jacob Devlin, Ming-Wei Chang, Kenton Lee, and Kristina Toutanova.
\newblock Bert: Pre-training of deep bidirectional transformers for language
  understanding.
\newblock {\em arXiv preprint arXiv:1810.04805}, 2018.

\bibitem[\protect\citeauthoryear{Duan \bgroup \em et al.\egroup
  }{2022}]{duan2022factorvae}
Yitong Duan, Lei Wang, Qizhong Zhang, and Jian Li.
\newblock Factorvae: A probabilistic dynamic factor model based on variational
  autoencoder for predicting cross-sectional stock returns.
\newblock In {\em Proceedings of the AAAI Conference on Artificial
  Intelligence}, volume~36, pages 4468--4476, 2022.

\bibitem[\protect\citeauthoryear{Fama and French}{1993}]{fama1993common}
Eugene~F Fama and Kenneth~R French.
\newblock Common risk factors in the returns on stocks and bonds.
\newblock {\em Journal of financial economics}, 33(1):3--56, 1993.

\bibitem[\protect\citeauthoryear{Fama and French}{2015}]{fama2015five}
Eugene~F Fama and Kenneth~R French.
\newblock A five-factor asset pricing model.
\newblock {\em Journal of financial economics}, 116(1):1--22, 2015.

\bibitem[\protect\citeauthoryear{Giglio \bgroup \em et al.\egroup
  }{2022}]{giglio2022factor}
Stefano Giglio, Bryan Kelly, and Dacheng Xiu.
\newblock Factor models, machine learning, and asset pricing.
\newblock {\em Annual Review of Financial Economics}, 14:337--368, 2022.

\bibitem[\protect\citeauthoryear{Gu \bgroup \em et al.\egroup
  }{2020}]{gu2020empirical}
Shihao Gu, Bryan Kelly, and Dacheng Xiu.
\newblock Empirical asset pricing via machine learning.
\newblock {\em The Review of Financial Studies}, 33(5):2223--2273, 2020.

\bibitem[\protect\citeauthoryear{Gupta \bgroup \em et al.\egroup
  }{2022}]{gupta2022hisa}
Ishu Gupta, Tarun~Kumar Madan, Sukhman Singh, and Ashutosh~Kumar Singh.
\newblock Hisa-smfm: historical and sentiment analysis based stock market
  forecasting model.
\newblock {\em arXiv preprint arXiv:2203.08143}, 2022.

\bibitem[\protect\citeauthoryear{Hu \bgroup \em et al.\egroup
  }{2018}]{hu2018listening}
Ziniu Hu, Weiqing Liu, Jiang Bian, Xuanzhe Liu, and Tie-Yan Liu.
\newblock Listening to chaotic whispers: A deep learning framework for
  news-oriented stock trend prediction.
\newblock In {\em Proceedings of the eleventh ACM international conference on
  web search and data mining}, pages 261--269, 2018.

\bibitem[\protect\citeauthoryear{Jensen \bgroup \em et al.\egroup
  }{1972}]{jensen1972capital}
Michael~C Jensen, Fischer Black, and Myron~S Scholes.
\newblock The capital asset pricing model: Some empirical tests.
\newblock 1972.

\bibitem[\protect\citeauthoryear{Ko and Lee}{2023}]{ko2023can}
Hyungjin Ko and Jaewook Lee.
\newblock Can chatgpt improve investment decision? from a portfolio management
  perspective.
\newblock {\em From a Portfolio Management Perspective}, 2023.

\bibitem[\protect\citeauthoryear{Li \bgroup \em et al.\egroup
  }{2022}]{li-etal-2022-csl}
Yudong Li, Yuqing Zhang, Zhe Zhao, Linlin Shen, Weijie Liu, Weiquan Mao, and
  Hui Zhang.
\newblock {CSL}: A large-scale {C}hinese scientific literature dataset.
\newblock In {\em Proceedings of the 29th International Conference on
  Computational Linguistics}, pages 3917--3923, Gyeongju, Republic of Korea,
  October 2022. International Committee on Computational Linguistics.

\bibitem[\protect\citeauthoryear{Liu \bgroup \em et al.\egroup
  }{2019}]{liu2019roberta}
Yinhan Liu, Myle Ott, Naman Goyal, Jingfei Du, Mandar Joshi, Danqi Chen, Omer
  Levy, Mike Lewis, Luke Zettlemoyer, and Veselin Stoyanov.
\newblock Roberta: A robustly optimized bert pretraining approach.
\newblock {\em arXiv preprint arXiv:1907.11692}, 2019.

\bibitem[\protect\citeauthoryear{Lopez-Lira and Tang}{2023}]{lopez2023can}
Alejandro Lopez-Lira and Yuehua Tang.
\newblock Can chatgpt forecast stock price movements? return predictability and
  large language models.
\newblock {\em arXiv preprint arXiv:2304.07619}, 2023.

\bibitem[\protect\citeauthoryear{Nakagawa \bgroup \em et al.\egroup
  }{2020}]{nakagawa2020ric}
Kei Nakagawa, Masaya Abe, and Junpei Komiyama.
\newblock Ric-nn: A robust transferable deep learning framework for
  cross-sectional investment strategy.
\newblock In {\em 2020 IEEE 7th International Conference on Data Science and
  Advanced Analytics (DSAA)}, pages 370--379. IEEE, 2020.

\bibitem[\protect\citeauthoryear{Schulman~J}{2017}]{ppo}
et~al. Schulman~J.
\newblock Proximal policy optimization algorithms.
\newblock {\em ArXiv}, 1707.06347, 2017.

\bibitem[\protect\citeauthoryear{Touvron \bgroup \em et al.\egroup
  }{2023}]{touvron2023llama}
Hugo Touvron, Thibaut Lavril, Gautier Izacard, Xavier Martinet, Marie-Anne
  Lachaux, Timoth{\'e}e Lacroix, Baptiste Rozi{\`e}re, Naman Goyal, Eric
  Hambro, Faisal Azhar, et~al.
\newblock Llama: Open and efficient foundation language models.
\newblock {\em arXiv preprint arXiv:2302.13971}, 2023.

\bibitem[\protect\citeauthoryear{Wu \bgroup \em et al.\egroup
  }{2022}]{wu2022s_i_lstm}
Shengting Wu, Yuling Liu, Ziran Zou, and Tien-Hsiung Weng.
\newblock S\_i\_lstm: stock price prediction based on multiple data sources and
  sentiment analysis.
\newblock {\em Connection Science}, 34(1):44--62, 2022.

\bibitem[\protect\citeauthoryear{Xie \bgroup \em et al.\egroup
  }{2023}]{xie2023wall}
Qianqian Xie, Weiguang Han, Yanzhao Lai, Min Peng, and Jimin Huang.
\newblock The wall street neophyte: A zero-shot analysis of chatgpt over
  multimodal stock movement prediction challenges.
\newblock {\em arXiv preprint arXiv:2304.05351}, 2023.

\bibitem[\protect\citeauthoryear{Xu \bgroup \em et al.\egroup
  }{2020}]{CLUECorpus2020}
Liang Xu, Xuanwei Zhang, and Qianqian Dong.
\newblock Cluecorpus2020: A large-scale chinese corpus for pre-training
  language model.
\newblock {\em ArXiv}, abs/2003.01355, 2020.

\bibitem[\protect\citeauthoryear{Yogo}{2006}]{yogo2006consumption}
Motohiro Yogo.
\newblock A consumption-based explanation of expected stock returns.
\newblock {\em The Journal of Finance}, 61(2):539--580, 2006.

\end{thebibliography}

\end{document}